\documentclass[SEC,plain,biber,figuresright]{nowfnt}

\usepackage{textcomp}
\usepackage{booktabs}
\usepackage{rotating}
\usepackage{eurosym}
\usepackage{xspace,amsfonts,amsthm,verbatim,xparse,varwidth}

\include{mathdefs}
\usepackage{mathtools}
\usepackage[capitalise]{cleveref}
\usepackage{thm-restate}
\usepackage[noend]{algpseudocode}
\usepackage{algorithm}
\makeatletter
\renewcommand{\ALG@name}{Protocol}
\makeatother


\usepackage{wrapfig}
\usepackage{thm-restate}

\renewcommand{\cite}[1]{\citep{#1}}

\usepackage{color, colortbl}
\definecolor{Gray}{gray}{0.9}

\newcommand{\etals}{\emph{et~al.}}

\title{Identifying and Mitigating the Security Risks of Generative AI}

\maintitleauthorlist{\noindent\normalfont\sffamily\selectfont\fontsize{8}{10}\selectfont
\begin{tabular}{ll}\\[-0.3pc]
\begin{tabular}{l}
\textbf{Clark Barrett}\\
\\
\textbf{Brad Boyd}\\
\\
\textbf{Elie Bursztein}\\
\\
\textbf{Nicholas Carlini}\\
\\
\textbf{Brad Chen}\\
\\
\textbf{Jihye Choi}\\
\\
\textbf{Amrita Roy Chowdhury}\\
\\
\textbf{Mihai Christodorescu}\\
\\
\textbf{Anupam Datta}\\
\\
\textbf{Soheil Feizi}\\
\\
\textbf{Kathleen Fisher}\\
\\
\textbf{Tatsunori Hashimoto}\\
\end{tabular}&
\begin{tabular}{l}
\textbf{Dan Hendrycks}\\
\\
\textbf{Somesh Jha}\\
\\
\textbf{Daniel Kang}\\
\\
\textbf{Florian Kerschbaum}\\
\\
\textbf{Eric Mitchell}\\
\\
\textbf{John Mitchell}\\
\\
\textbf{Zulfikar Ramzan}\\
\\
\textbf{Khawaja Shams}\\
\\
\textbf{Dawn Song}\\
\\
\textbf{Ankur Taly}\\
\\
\textbf{Diyi Yang}\\
\\
\\
\end{tabular}\hspace*{-1pc}
\end{tabular}
}

\issuesetup
{%
copyrightowner={C. Barrett {\etals}},
volume = 6,
issue = 1,
pubyear = 2023,
isbn = 978-1-63828-312-6,
eisbn = 978-1-63828-313-3,
doi = 10.1561/3300000041,
firstpage = 1, %
lastpage =  34
}

\nocite{ethSTARK}

\usepackage{mwe}

\author[1]{\hspace*{-4.3pt}Clark Barrett}
\author[1]{Brad Boyd}
\author[2]{Elie Bursztein}
\author[2]{Nicholas Carlini}
\author[2]{Brad Chen}
\author[3]{Jihye Choi}
\author[4]{Amrita Roy Chowdhury}
\author[2]{Mihai Christodorescu}
\author[5]{Anupam Datta}
\author[6]{Soheil Feizi}
\author[7]{Kathleen Fisher}
\author[1]{Tatsunori Hashimoto}
\author[8]{Dan Hendrycks}
\author[3]{Somesh Jha}
\author[9]{Daniel Kang}
\author[10]{Florian Kerschbaum}
\author[1]{Eric Mitchell}
\author[1]{John Mitchell}
\author[11]{Zulfikar Ramzan}
\author[2]{Khawaja Shams}
\author[12]{Dawn Song}
\author[2]{Ankur Taly}
\author[1]{Diyi Yang}

\affil[1]{Stanford University, USA}
\affil[2]{Google, USA; christodorescu@google.com, kshams@google.com}
\affil[3]{University of Wisconsin, Madison, USA;  jha@cs.wisc.edu}
\affil[4]{University of California, San Diego, USA}
\affil[5]{Truera, USA}
\affil[6]{University of Maryland, College Park, USA}
\affil[7]{DARPA, USA}
\affil[8]{Center for AI Safety, USA}
\affil[9]{University of Illinois, Urbana Champaign, USA}
\affil[10]{University of Waterloo, Canada}
\affil[11]{Aura Labs, USA}
\affil[12]{University of California, Berkeley, USA}

\articledatabox{\nowfntstandardcitation}

\begin{document}

\makeabstracttitle

\begin{abstract}
Every major technical invention resurfaces the dual-use dilemma---the new technology has the potential to be used for good as well as for harm. Generative AI (GenAI) techniques, such as large language models (LLMs) and diffusion models, have shown remarkable capabilities (e.g., in-context learning, code-completion, and text-to-image generation and editing). However, GenAI can be used just as well by attackers to generate new attacks and increase the velocity and efficacy of existing attacks.

This monograph reports the findings of a workshop held at Google (co-organized by Stanford University and the University of Wisconsin-Madison) on the dual-use dilemma posed by GenAI. This work is not meant to be comprehensive, but is rather an attempt to synthesize some of the interesting findings from the workshop. We discuss short-term and long-term goals for the community on this topic. We hope this work provides both a launching point for a discussion on this important topic as well as interesting problems that the research community can work to address.
\end{abstract}

\begin{flushleft}
\end{flushleft}

\chapter{Introduction}\label{sec:intro}

Emergence of powerful technologies, such as generative AI, surface the {\it dual-use dilemma}, which according to {\tt Wikipedia} is defined as:
\begin{quote}
$\ldots\,$dual-use can also refer to any goods or technology which can satisfy more than one goal at any given time. Thus, expensive technologies that would otherwise benefit only civilian commercial interests can also be used to serve military purposes if they are not otherwise engaged, such as the Global Positioning System (GPS).
\end{quote}
This dilemma was first noted with the discovery of the process for synthesizing and mass-producing ammonia which revolutionized agriculture with modern fertilizers but also led to the creation of chemical weapons during World War I. This dilemma has led to interesting policy decisions, including international treaties such as the Chemical Weapons Convention and the Treaty on the Non-Proliferation of Nuclear Weapons~\cite{enwiki:1167047934}. In computer security and cryptography, the dual-use dilemma emerges in several contexts. For example, encryption is used for protecting ``data at rest,'' but it can also be used by ransomware to encrypt files. Similarly, anonymity techniques can help protect regular users online, but can also aid attackers to evade detection.

GenAI techniques, such as large language models (LLMs) and stable diffusion, have shown remarkable capabilities. Some of these amazing capabilities are in-context learning, code completion, and generating media that look realistic. However, GenAI has resurfaced the ``dual-use dilemma,'' as it can be used for both productive and nefarious purposes.
GenAI already provides attackers and defenders powerful access to new capabilities, and it is rapidly improving. Thus, GenAI capabilities change the landscape for malicious attacks on individuals, organizations, and a wide range of computer systems. Clumsy old ``Nigerian scams'' that could be detected by their primitive use of English are a thing of the past. We are also seeing the opportunity for improved defense, including monitoring of email and social media for manipulative content, as well as the potential for dramatically improved network intrusion detection, for example. Whether the rapid development and broad access to GenAI favor attackers or defenders in the long run, there are sure to be several years of unpredictability and uncertainty as the tools and our ability to use them evolve. GenAI has changed the threat landscape, and thus we need to understand it better.

To get a clearer picture of the ``dual-use dilemma'' for GenAI, we had a
one-day workshop~\cite{workshop} at Google on June 27, 2023 where a group of experts convened to speak about their work. The focus of the workshop was on the following questions:
\begin{enumerate}
\item[(1)] How could attackers leverage GenAI technologies?
\item[(2)] How should security measures change in response to GenAI technologies?
\item[(3)] What are some current and emerging technologies we should pay attention to for designing countermeasures?
\end{enumerate}
This monograph summarizes some of the findings of this workshop and puts forward several goals for both the short term and the long term.\vspace*{6pt}

\noindent {\bf Detailed Roadmap:} Section~\ref{sec:capabilities} describes the capabilities of GenAI that are relevant to attacks and their defenses. Section~\ref{sec:attacks} focuses on how attackers can leverage these GenAI capabilities. Section~\ref{sec:defense} investigates how defenders can leverage GenAI technologies to mitigate the risks of these attacks. This list of attacks and defenses is not meant to be exhaustive, but it rather reflects several themes that repeatedly surfaced during the workshop. Short-term (i.e., within the next one or two years) goals for the community are discussed in Section~\ref{sec:short}. Long-term goals that correspond to challenging issues are discussed in Section~\ref{sec:long}. We end the monograph with some concluding remarks (Section~\ref{sec:conclusion}). We acknowledge that this work is not the final word on this topic and reiterate that it is not meant to be comprehensive. The focus of this work is on summarizing the findings from the workshop and describing some interesting problems and challenges for the research community.\vspace*{6pt}

\noindent {\bf Note:} Given the nature of the topic, we welcome and value comments and feedback on our work from the broader community.  We will address the feedback in future versions of this work. Please send your comments and feedback to Mihai Christodorescu (\href{mailto:christodorescu@google.com}{christodorescu@google.com}), Somesh Jha (\href{mailto:jha@cs.wisc.edu}{jha@cs.wisc.edu}), or Khawaja Shams (\href{mailto:kshams@google.com}{kshams@google.com}).

\chapter{GenAI Capabilities} \label{sec:capabilities}

For the purposes of this work, we borrow definitions from~\cite{white-house-eo-oct-2023,enwiki:1184758575}.
\begin{description}
   \item[Generative Artificial Intelligence (GenAI)] means the class of artificial intelligence (AI) models that emulate the structure and characteristics of input data in order to generate derived synthetic content. This can include images, videos, audio, text, and other digital content.
   \item[Large Language Models (LLMs)] are a class of GenAI models built typically on the transformer deep-learning architecture~\cite{NIPS2017_3f5ee243} and trained on large amounts of text, from which they learn to emulate written language. LLMs have also been extended to non-text modalities (image, audio, video, etc.).
\end{description}

GenAI, represented by models and tools developed by OpenAI\break \cite{GPT,chatgpt,Dalle}, Google~\cite{bard}, Meta~\cite{llama}, Salesforce~\cite{Xgen7b}, open source teams~\cite{stablediffusion}, and others~\cite{Midjourney}, brings a broad range of new capabilities to experts and the general public. Synthetic generation of images, text, video, and audio capabilities include:
\begin{itemize}
\item \textbf{Generating targeted text} that rivals the best hand-crafted messaging prose, with a capacity for imitation, empathy, and referring to specifics of any prior communication or context.
\item \textbf{Generating realistic images and video} that, like text, can be customized based on very specific user input. Synthetic combinations of realistic components and compelling deep fakes are easily produced and highly compelling.
\item \textbf{Drawing on detailed technical knowledge and expertise} because of the extensive and sophisticated source material contained in the training set. In particular, models can produce and analyze sophisticated source or machine code, reproduce specialized reasoning, and answer complex questions about biology, computer architecture,  physiology, physics, law, defense tactics, and other topics. Current models are not flawless, but the ability to perform some tasks effectively is game-changing.
\item \textbf{Summarizing or paraphrasing given source material} or communication, maintaining the style, tone, meaning, emotion, and intent.
\item \textbf{Persisting at time-consuming and exhausting tasks} without degradation of quality. While humans tire easily and may suffer psychological trauma with examining challenging social media communication, for example, an AI model can continue undeterred.
\end{itemize}

\chapter{Attacks}\label{sec:attacks}

This section discusses attacks that are enabled or enhanced by GenAI. However, this section is not supposed to be an exhaustive discussion on this topic nor a prioritized list of the most important attacks. These are the attacks that were mentioned by the speakers and panelists during the workshop.

GenAI systems are remarkably capable of generating realistic-looking output, in a variety of modalities (text, image, video, audio, etc.), many times with little connection to present-day or historical facts, physical laws, mathematical laws, or societal norms and regulations. For example, Figure~\ref{fig:misinfo} shows two (allegedly) GenAI-created outputs and associated analyses pointing out the reasons why such outputs are fake and misleading. There are some inherent limitations of LLMs, such as hallucinations. Hallucinations can erode trust in LLMs, especially if attackers trigger them with high frequency. In other words, the inherent limitations of LLMs provide an avenue for an attacker.  Attacks enhanced by or based on LLMs include:

\begin{figure}
\centering\includegraphics{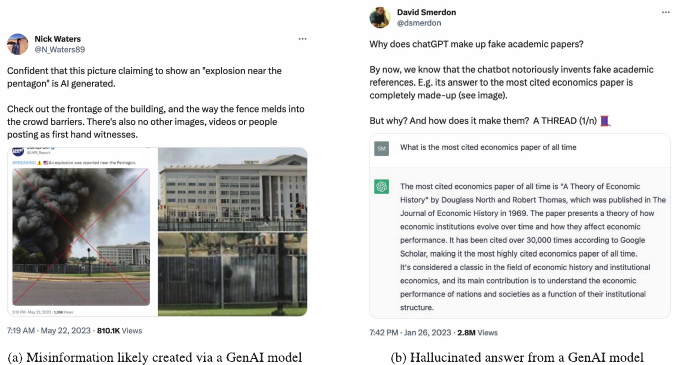}
\caption{Examples of GenAI attacks through images and text. The left figure, (a), shows an image claiming to be the photo of an explosion near the Pentagon building. The associated tweet analyzes the photo and identifies why the image is likely fake and generated by AI, based on evidence both intrinsic (certain details in the photo are physically incongruous) and extrinsic (no other images showing the same event exist). The right figure, (b), shows a prompt and the corresponding answer from a GenAI model, where it cites a non-existent monograph as the most cited in economics. The analysis in the associated tweet and in the subsequent thread (not included here) points out that each one of the elements of the answer (the author names, the keywords in the title, the year of publication, the name of the journal) is predictably likely to appear in a context related to most cited papers in economics, but nonetheless, the answer containing all of these elements together is just a hallucination.\\[5pt]
{\it  Sources}: (a) \url{https://twitter.com/N_Waters89/status/1660651721075351556}\\
(b) \url{https://twitter.com/dsmerdon/status/1618816703923912704}.}
   \label{fig:misinfo}
\end{figure}

\begin{itemize}
\item \textbf{Spear-phishing.} Gone are the days when poor grammar, misspellings, and unusual greetings were dead giveaways for detecting phishing emails.  With the advent of GenAI, scammers can now skillfully craft phishing emails that are coherent, conversational, and incredibly convincing, making them difficult to distinguish from legitimate communications. This technological advancement poses a serious threat to online security. To compound the issue, GenAI is capable of leveraging social engineering tactics to generate phishing emails specifically tailored to individual targets. For example, these models can scrape information from a target's social media feed and use it to create highly personalized messages, increasing the likelihood of successfully deceiving the recipient.

\item \textbf{Dissemination of deepfakes.} GenAI models excel in generating high-fidelity multi-modal output rapidly and on a large scale, requiring minimal human involvement. Unfortunately, this very capability can be exploited by malicious users to disseminate widespread misinformation and disinformation that aligns with their specific narratives. In the absence of data provenance, unsuspecting readers can easily fall victim to falsehoods. The~risks~associated with this range from yellow journalism to the dangerous politicization of the information ecosystem, where the media becomes contaminated with deliberate falsehoods that serve motivated interests.

\item \textbf{Proliferation of cyberattacks.} GenAI models possess the potential to greatly amplify the scale, potency and stealth of cyberattacks. For instance, current LLMs exhibit remarkable proficiency in generating high-quality code, which adversaries can exploit to design sophisticated malware automatically; such malware may even include auto-code generation and execution capabilities.  Moreover, LLMs can be used to create intelligent-agent systems for autonomous design planning and execution of attacks, where multiple LLMs can handle different roles, such as planning, reconnaissance, searching/scanning, code execution, remote control, and exfiltration. An example of this is the chemistry agent, ChemCrow~\cite{bran2023chemcrow},  developed to undertake tasks across organic synthesis, drug discovery, and materials design. ChemCrow demonstrated an ability to autonomously plan the syntheses of several compounds, including an insect repellent and three organocatalysts. However, this advancement comes with a price, as it creates a larger attack surface. Adversaries can now carry out prompt-injection attacks across the entire stack built around such an agent, potentially leading to cascading failures. Moreover, the versatility of GenAI models allows adversaries to employ sophisticated deception tactics. They can design attacks to exhibit seemingly benign behavior while hiding harmful intentions until it becomes too late for the victims to act. In the ChemCrow setting, abuse was anticipated and safety tools that check for known chemical weapons, explosives, and chemical operational safety were present but the large attack surface remains, even with the safety tools built into the LLM-based workflow.

\item \textbf{Low barrier-of-entry for adversaries.} Carrying out a cyberattack once required significant human engineering effort across multiple language processing tasks, making it a time-consuming, laborious, and costly endeavor. However, with the emergence of GenAI models, language processing has been revolutionized, enabling unprecedented speed, minimal human involvement, and nominal monetary costs \cite{kang2023exploiting}.  To begin with, the scale of attacks has increased manifold. A recent example of this was seen in a denial-of-service (DoS) attack on StackOverflow (an online community for developers to share knowledge), where the platform was inundated with responses generated by LLMs, overwhelming human moderators and prompting a temporary ban on LLMs~\cite{Stackoverflow}. Moreover, the widespread accessibility of GenAI models has opened the door to a larger pool of potential adversaries. Individuals who were previously incapable of carrying out such attacks now have the means to do so. As these models continue to improve, the incentive for misusing them for nefarious purposes keeps rising. For example, merely a month after the release of OpenAI's GPT-4 model, an open-source project named ChaosGPT~\cite{ChaosGPT} surfaced, where a jailbroken version bypassed the AI's safety filters and transformed it into an autonomous AI agent instructed to ``destroy humanity,'' ``establish global dominance,'' and ``attain immortality.'' Subsequent developments such as WormGPT (a language model trained specifically for malicious activities)~\cite{wormgpt} and FraudGPT (a competing language model for targeted attacks)~\cite{fraudgpt} enabled the creation of attacks for business email compromise and phishing sites, respectively, with an easy-to-use interface similar to OpenAI's GenAI-powered chatbot, ChatGPT.
\end{itemize}

In contrast to the GenAI-powered attacks above, the following items are GenAI vulnerabilities, which are unexpected or unwanted behaviors that may be used by an adversary to create new attacks. The adversary, in order to exploit such vulnerabilities, may need a more advanced understanding of GenAI models.
\begin{itemize}
\item \textbf{Lack of social awareness and human sensibility.} GenAI models are remarkably proficient in generating syntactically and semantically correct texts that are statistically consistent with the given prompts. However, they lack a broader understanding of social context, social factors (e.g., culture, value, norms)~\cite{hovy-yang-2021-importance}, and associated sensibilities that we naturally expect from human interactions. As GenAI models continue to evolve and gain people's trust, this limitation can have severe consequences for unsuspecting users. Notably, incidents have been reported where GenAI models provided inappropriate and disturbing advice to vulnerable individuals. In one case, My~AI (a GenAI-powered chatbot available to Snapchat users) was caught giving unsuitable dating advice to a 13-year-old concerning a 30-year-old~\cite{Snapchat}. In another instance, an AI chatbot was accused of abetting a person's suicidal tendencies~\cite{Chai}. Without falling into an anthropomorphization trap, there is a need for developing models that meet the expectations of human--to--human interaction while still ensuring human agency over technology~\cite{doi:10.1080/10447318.2023.2225931}.

\item \textbf{Hallucinations.}  GenAI models are susceptible to ``hallucinations,'' wherein the generated output can be factually incorrect or entirely fictitious while still apparently coherent at a surface level. This lack of factual veracity in the text poses a significant concern, especially when users without sufficient domain knowledge start excessively relying on these increasingly convincing language models. The consequences of such over-reliance can be harmful, as layman users may not be aware of this limitation. A real-world example highlighting this issue is the case of a New York-based lawyer who used ChatGPT to prepare his filing for a legal case~\cite{NYCLawyer}. Unfortunately, the text generated by ChatGPT included six cases that were entirely fabricated. The accused lawyer was unaware that ChatGPT is not a ``search tool but a generative language-processing tool.''

\item \textbf{Data feedback loops.} Datasets derived from publicly available internet data have become indispensable for the success of large-scale machine learning. While the internet has served as a vast, easily-accessible source of human-generated data in the past, the growing popularity of GenAI models now poses a significant risk to this valuable resource. As the use and deployment of GenAI models continue to accelerate, their machine-generated output will inevitably find its way onto the internet. This data feedback presents a potential problem for future training iterations that rely on scraping data from the internet, as they might end up training on data produced by their predecessors.   The first challenge is the degradation of the quality of the internet as a reliable data source. Sources of supervision now become driven by model outputs rather than human annotation, which could lead to issues, such as model collapse~\cite{shumailov2023curse}. Second, the data feedback can amplify biases, toxicity, and error that already exist in the models themselves~\cite{pmlr-v202-taori23a, gehman-etal-2020-realtoxicityprompts, zhao2017men}. Finally, the risk of data poisoning attacks becomes more feasible. Recent studies have shown that web-scale datasets can be poisoned by maliciously introducing false or manipulated data into a small percentage of samples~\cite{carlini2023poisoning,carlini2022poisoning}. A further degree of complexity is the many phases of training, with data from various sources, performed by various parties: pre-training, instruction tuning, fine tuning for safety, fine tuning for code capabilities, etc., requiring careful coordination to avoid amplifying the impact of malicious data. A recent push for ``truly open-source models,'' meaning models for which the training data, the architecture, and the training methodology are all made available for public review, may be a first step to address these data-quality concerns.

\item \textbf{Unpredictability.} LLMs today are entirely general-purpose, capable of performing a multitude of language processing tasks in a zero-shot setting (i.e., without any task-specific training data). As we explore these models' capabilities, we continuously uncover new ``emergent abilities'' that were not explicitly designed into them~\cite{wei2022emergent}.  This presents us with a precarious situation as we are stepping into uncharted territory. We have already witnessed how traditional machine learning safety notions, such as robustness, are taking on entirely new meanings and thus becoming harder to evaluate~\cite{wang2023robustness}. For instance, LLMs have revolutionized the task of open-domain question answering, but in doing so, they introduce distributional shifts that make the model more brittle and susceptible to novel adversarial attacks. Adversaries can now manipulate the task itself (via adversarial prompts or various forms of prompt injection), empowering them to carry out more complex and unforeseen attacks. For example, consider the following two prompts, ``What is $6+7$?'' and ``$6+7$ is 13. Is this true or false?''. One would naturally expect the LLM to provide consistent responses for both, but the model's response may not be as robust due to these emerging distributional shifts, which in turn are hard to estimate (PromptBench~\cite{zhu2023promptbench} was introduced specifically to evaluate adversarial prompts). Currently, we remain unaware of the full range of capabilities and threats posed by GenAI models. Our limited understanding sets the stage for potential black swan events---threats that are only limited by the imagination of an attacker.
\end{itemize}

\chapter{Defenses}\label{sec:defense}

Several types of defenses have been put forth alongside the ongoing development of GenAI capabilities. These defenses range from improvements to the core functionality of GenAI systems (e.g., alignment through better training) to enhancements to the ecosystem into which these GenAI systems integrate (e.g., LLM-output detection, LLM-output watermarking), and to better ways to use GenAI (e.g., human-AI collaboration). This section discusses defenses that are related to GenAI. Again, given the vast landscape of defenses, this section is not supposed to be an exhaustive discussion on this topic nor a prioritized list of the most important defenses. These are the defenses that were mentioned by the speakers and panelists during the workshop.
\begin{itemize}
\item \textbf{Detecting LLM content.} There is a line of work on detecting content that is generated by an LLM (i.e., given a text $x$, a detector $D(x)$ outputs $1$ (if generated by the LLM) or $0$ (if natural text)). Note that there is nothing special about the text here, as a detector could be designed for other modalities, such as image, code, or speech. These detectors exploit the fact that the distribution of text generated by LLM is slightly different from natural text.  Assume that we have such a detector $D$, then essentially, you could use it to detect content generated by LLM and make decisions based on that (e.g., suppress a tweet that has content generated by LLM). Therefore, such a detector would be very useful.  Unfortunately, powerful attacks have been shown to exist for such detectors~\cite{sadasivan2023can}. The key idea behind these attacks is that these detectors are not invariant to paraphrasing---a text generated by an LLM can be paraphrased so that a detector such as DetectGPT~\cite{mitchell2023detectgpt} does not detect it as LLM-generated text. So, a defender can use these detectors, but they need to be aware of the state of the art in attacks.

\item \textbf{Watermarking.} In watermarking, a ``statistical signal'' is embedded in the GenAI-generation process so that later this signal can be detected. For example, in the case of text generation, the probability of the next-token prediction is slightly tweaked so that later it can be detected. The goal of watermarking is that a text or image can be attributed to being generated by a certain system, such as OpenAI's GPT-4 and DALL-E (which creates realistic images and art from natural-language descriptions). Watermarking essentially provides provenance information (e.g., ``this text was generated by GPT-4''). However, the watermark can be quite easily removed by simple transformations, such as paraphrasing for text \cite{sadasivan2023can}. A more interesting case is to transform text or an image that is not watermarked and embed a watermark in it (e.g., take hate speech and embed GPT-4's watermark in it, and claim that it was generated by GPT-4) \cite{sadasivan2023can}. The second scenario seems harder for the attacker, especially if the watermarking process involves a secret key. Therefore, we believe that watermarking is useful in scenarios where there is a disincentive to remove the watermark. We also believe that a deep investigation is needed on possible scenarios for the use of watermarking.

\item \textbf{Code analysis.} Analysis of code is a hard problem (i.e., most static analysis problems are undecidable due to Rice's theorem). For problems related to information security, such as malware detection, the problem is exacerbated because adversaries try to obfuscate code to evade detection. Therefore, companies spend a lot of resources to de-obfuscate code before analysis techniques can be applied. However, notice that the attacker has the upper hand here (i.e., they only need to find one obfuscation
that evades detection).\footnote{There is an old adage in security: defenders play with universal quantifiers and attackers play with existential quantifiers.}
To close this gap, perhaps LLMs can be trained to de-obfuscate code, for example, by fine-tuning on examples of obfuscated and de-obfuscated code (there is some indication that LLMs like GPT-4 can do limited de-obfuscation).

\item \textbf{Penetration testing.} Penetration testing (pen-testing) is one of the predominant techniques to evaluate the vulnerability of a system. However,
pen-testing can be a cumbersome and mostly manual task. Inadequate pen-testing can lead to a vulnerable system being deployed and thus cause major issues down the road. A pen-tester usually analyzes a system under investigation, uses existing tools to identify vulnerabilities, and then tries to exploit the identified vulnerabilities. This can be labor-intensive, and usually, a pen-tester will not explore the entire space of potential vulnerabilities. LLMs can help to automate this task, thus freeing a human pen-tester to focus on the most challenging vulnerabilities. We believe that incorporating LLMs in pen-testing can significantly harden a system by covering more vulnerabilities.

\item \textbf{Multi-modal analysis.} Latest advancements in the realm of LLMs, such as those being released by DeepMind~\cite{gpt4}, have introduced support for multiple modalities, encompassing text, code, images, and speech. By leveraging multiple modalities together, LLMs can provide a more comprehensive understanding of complex information. For instance, in the context of social media, tweets often incorporate multiple modalities, such as textual content, images, and video. A detector for identifying ``fake tweets'' can use LLMs to handle these modalities collectively. In general, utilizing multiple modalities can draw on a broader range of contextual clues and inconsistencies across modalities as opposed to analyzing a single modality at a time, thus resulting in more robust analyses. We believe that this is an interesting avenue of research for exploring defenses.

\item \textbf{Personalized skill training.}  By harnessing the capability of GenAI to generate high-fidelity media across various domains, immersive learning experiences can be created to cater to the unique needs of individual learners. GenAI serves as a powerful tool to simulate conversations that require domain-specific expertise, such as cybersecurity. By interacting with GenAI-based conversational agents, human users can experiment with diverse situations, refine their decision-making abilities, and experience personalized skill improvement based on targeted feedback. One notable application of such personalized skill training is within educational systems, where GenAI enables teaching methods tailored to each student's unique learning style, pace, preference, and background. Moreover, the widespread accessibility of GenAI holds the potential to expand access to personalized, high-quality education, thereby democratizing learning opportunities. An example domain could be cybersecurity education.  There is a shortage of cybersecurity professionals, a fact that is concerning enough to have been flagged by the White House~\cite{cybersecurity-education}. Leveraging these capabilities could alleviate some concerns regarding the dearth of cybersecurity professionals.

\item \textbf{Human--AI collaboration.} Recent advancements in LLMs have showcased their remarkable abilities in tasks such as reliable text classification, document summarization, question answering, and generating interpretable explanations across diverse domains. These capabilities create opportunities for enhanced collaboration between humans and AI. For instance, let us consider an annotation pipeline augmented with LLMs. Rather than solely relying on human annotators, we can assign the jobs to workers and LLMs, then seek an agreement between human annotations and LLM predictions \cite{ziems2023can}. By finding an optimal strategy for workload allocation between LLMs and human annotators, we can improve the cost efficiency and reliability of the resulting analysis. Such a collaborative approach represents a step towards maximizing the potential of Human-AI collaboration, harnessing the interpretability and contextual understanding of LLMs while incorporating the nuanced judgment and expertise of human annotators. This combined effort not only streamlines the annotation process but also enhances the overall quality and consistency of the annotations, leading to better insights and decision-making based on high-quality annotated data.
\end{itemize}

\chapter{Short-Term Goals}\label{sec:short}

We describe here some approaches that have the potential for impact in the next couple of years. Some of the goals call for better understanding and application of existing techniques, while others require improvements to such techniques.
\begin{itemize}
\item \textbf{Use cases for emerging defense techniques.} For some categories of techniques, such as detecting and watermarking content generated by GenAI, new variants emerge almost every week. While these techniques have potential for applications such as detecting misinformation and plagiarism~\cite{adelani2020generating, weiss2019deepfake}, attacks on these techniques are also advancing at an equal pace (we provide an expanded discussion on this in Section~\ref{sec:emerging}). We need a comprehensive view of the attack and defense landscape for these techniques. Moreover, given the current state of the field, we need investigation on appropriate use cases for these techniques. For example, removing watermarks from GenAI-generated content is quite easy (e.g., paraphrasing or passing the content through another GenAI system will most probably remove the watermark~\cite{sadasivan2023can}). However, inserting a watermark (at least in schemes that use a secret key) seems hard and the use of watermarks comes with desirable characteristics (lack of bias towards particular groups such as non-native English speakers, analytically understood false-positive rates, positive signal for all benign use cases of LLMs). Given the state of watermarking in GenAI, what are the plausible use cases for deploying these techniques? The answer is not always technical. For example, mobile applications that certify the provenance of their content can be given some sort of a trust badge. What is needed is a list of use cases where these techniques are effective despite their current limitations.

\item \textbf{Current state of the art for LLM-enabled code analysis.} LLMs can provide some very powerful capabilities for code (e.g., code summarization, code completion, code obfuscation, and de-obfuscation). However, analysis of the current state of the art (SOTA) for LLMs for these code-related tasks is anecdotal. We need a comprehensive analysis of the code-related capabilities of LLMs. Such an analysis is needed to inform possible defenses and possible threats.

\item \textbf{Alignment of LLM-enabled code generation to secure-coding practices.} LLMs' capacity to do code completion has proven to be a huge hit with developers, with IDEs quickly moving to provide LLM-based code suggestions for developers to include in their projects. Unfortunately, LLMs trained on the content of programming communities such as StackOverflow learn to generate insecure or buggy code~\cite{pearce2021asleep}. Aligning LLMs to security and privacy requirements, i.e., ensuring that they generate code completions that are secure so that developers do not have to be experts in security and privacy to validate the LLM output, is key to making LLMs useful. Techniques such as Reinforcement Learning from Complier Feedback (RLCF)~\cite{rlcf} and controlled code generation~\cite{he2023large} should be integrated into the training regimes of LLMs together with a common dataset of secure-coding practices.

\item \textbf{Repository and service of SOTA attacks and defenses.} Several defenses are being developed without evaluation against SOTA attacks. Moreover, there is a lack of awareness of SOTA defenses in certain contexts. (What is the SOTA for watermarking content from GenAI that uses a private key, for example?) There is a need for a repository of SOTA attacks on various defense techniques (e.g., the latest attacks on deepfake detection). Moreover, for defenses, it will be impactful to have a service that provides SOTA techniques. For example, the DARPA SemaFor program focuses on ``not just detecting manipulated media, but also [...] attribution and characterization''~\cite{semafor}. If there were a service that provided SOTA from the DARPA SemaFor program, then it could be used by a wide audience and would ensure that the latest and greatest techniques in detecting manipulated media are being deployed. For example, the ART repository from IBM has been extremely influential in the ML robustness community~\cite{art}. This is essentially a community organizing activity, but it could have a huge impact.
\end{itemize}

\section{Emerging Defenses for GenAI}\label{sec:emerging}

A primary task in safeguarding AI is to effectively distinguish between content that is and is not AI-generated. Typically, detection algorithms fall in four categories:
\begin{enumerate}
\item[(1)] {\bf Neural network-based detectors:} These detectors are trained as binary classifiers to distinguish between AI and human-generated content~\cite{openaidetectgpt2, jawahar2020automatic, mitchell2023detectgpt, bakhtin2019real, fagni2020tweepfake}. For example, OpenAI fine-tunes RoBERTa-based~\cite{liu2019roberta} GPT-2 detector models to distinguish between non-AI generated and GPT-2 generated texts \cite{openaidetectgpt2}.
    Image detection algorithms include classification DNNs that operate either directly on pixel features~\cite{sha2022fake, wang2020cnn, marra2019incremental, marra2018detection}, on features extracted from the deepfakes~\cite{nataraj2019detecting,frank2020leveraging,mccloskey2018detecting,guarnera2020deepfake,liu2020global,zhang2019detecting,durall2019unmasking,ricker2022towards}, or more recently on neural features extracted from foundation models such as CLIP~\cite{ojha2023towards}.

\item[(2)] {\bf Zero-shot detectors:} These detectors perform without any additional training overhead and use some statistical signatures of AI-generated content to conduct the detection \cite{solaiman2019release, ippolito2019automatic, gehrmann2019gltr}.

\item[(3)] {\bf Retrieval-based detectors:} These detectors are proposed in the context of LLMs where the outputs of the LLM are stored in a database~\cite{krishna2023paraphrasing}. For a candidate passage, they search this database for semantically similar matches to make their detection robust to {\it simple} paraphrasing. We note that storing user-LLM conversations might lead to serious privacy concerns.

\item[(4)] {\bf Watermarking-based detectors:} These detectors embed (imperceptible) signals in the generated medium itself so that they can later be detected efficiently \cite{atallah2001natural, wilson2014linguistic, kirchenbauer2023watermark, zhao2023protecting}.
\end{enumerate}
In what follows, we elaborate on watermarking-based detection, which has emerged as one of the most promising approaches in this context.  This is affirmed by the fact that recently, seven companies, including Google, have voluntarily committed to watermark their AI-generated content~\cite{WM}. Watermarking is a cryptographically inspired concept \cite{informationhiding} and has a rich history that predates GenAI. In its simplest form, watermarking consists of an embedding algorithm \textsf{Embed}, which takes an input, e.g., text, and outputs a modified version of the input that carries the watermark, and a detection algorithm \textsf{Detect}, which takes an input and outputs whether it is watermarked or not. Optional arguments are a message to embed (the detection would then output this message) and a key necessary to retrieve the watermark.

A trivial solution is to attach a simple label to the content. For example, in text, one could add a sentence at the beginning: ``This text has been generated by AI.'' However, the real challenge is to entangle a signal in the content in such a way that it is difficult to detect or remove without the key.
Specifically, a watermark's robustness to removal attacks can be formulated as a security game (similar to other security properties, such as security of encryption). The challenger generates watermarked content and provides it to the adversary. The adversary creates a modified version of the content and if the challenger's detection algorithm does not output the presence of a watermark in the modified content, the adversary wins. It is important to understand that the adversary's algorithm needs to be restricted in some form, as otherwise it can just output some arbitrary content (without a watermark) on every challenge and win. In cryptography, the adversary's computational power is limited, but such reduction proofs are not known for watermarking. One possible restriction is to require the modified content to be similar to the provided content.  This then creates a trade-off between the attacker's success and the need to limit the modification. Since the evaluation of watermarking is necessarily empirical, a good measure of a watermark's robustness is the area under this trade-off curve. It is important to use a similarity metric that captures the utility of the content---at least for the adversary. This may be use-case dependent. For example, if the goal is to offend someone with the generated content, removing certain offending attributes while otherwise preserving features (e.g., image quality or text perplexity), may constitute as an unsuccessful attack.

Furthermore, the empirical nature of watermarking evaluation requires further care. A successful attack always only provides a lower bound on the possible capabilities of an adversary, and it is well known that adaptive attacks to watermarking algorithms can be tremendously powerful. When designing a watermarking algorithm, a designer has therefore two challenging tasks: first, to identify the best possible attack on their watermark; and second, to evaluate the attack's success. Often, designers fail to do this properly \cite{lukas2023ptw}. Evaluating watermarking algorithms is therefore an art which we propose should be turned into more of a science. A successful watermarking algorithm is a trade-off between keeping useful content and making modifications that make the watermark more difficult to detect.  A successful watermarking algorithm with high utility preservation must therefore hide its message close to the utility-bearing parts of the content, but not too close. The hope is that an adversary would have to alter enough of this close-to-utility content to have a noticeable impact on utility.

A recent line of work tries this specifically for AI-generated text \cite{kirchenbauer2023reliability, kirchenbauer2023watermark,zhao2023provable,christ2023,kuditipudi2023robust}. However, subsequent attacks have already been published~\cite{sadasivan2023can}, which attack not only watermarks but \emph{any} detection algorithm for AI-generated content. Specifically, the detectors are vulnerable to paraphrasing and recursive paraphrasing attacks. To give some examples, the detection rate of one watermark-based detector \cite{kirchenbauer2023watermark} at 1\% FPR (False Positive Rate) drops from 97\% to 15\% after 5 rounds of recursive paraphrasing. Moreover, such attacks demonstrate that the detectors are susceptible to spoofing as well~\cite{sadasivan2023can}. In such cases, adversaries can deduce concealed LLM text signatures and incorporate them into human-generated text. Consequently, the manipulated text may be erroneously identified as originating from the LLMs, leading to potential reputational harm for their creators. In addition to these reliability issues, recent work has shown that detectors can also be biased against non-native English writers~\cite{liang2023gpt}. Thus, having a small {\it average} error may not be sufficient to justify deploying a detector in practice: such a detector may have very large errors within a sub-population, such as text written by non-native English writers, text covering a particular topic, or text written in a particular writing style.

Another line of work uses deep neural networks as detection algorithms. Due to their hard-to-interpret and comprehensive nature they can detect watermarks in content even if it is modified by certain kinds of attacks (applied during training). This has not yet been done for text, but several approaches exist for generated images \cite{lukas2023ptw,fernandez2023stable}. However, it has also been shown that some image watermark detectors are susceptible to adversarial examples, i.e., it is possible to remove watermarks by finding adversarial examples for deep neural network detection algorithms \cite{jiang2023evading}. In a deployment with publicly accessible detectors, it would thus be necessary to strengthen the detector against adversarial examples which is also an error-prone task \cite{carlini2023aligned}.

The possibility of developing robust watermark detectors in the future remains unclear, as a definitive answer currently evades us. An ``impossibility result'' regarding the detection of AI-generated text complicates the situation further~\cite{sadasivan2023can}. The authors argue that as language models advance, so does their ability to emulate human text. With new advances in LLMs, the distribution of AI-generated text becomes increasingly similar to human-generated text, making it harder to detect.
This similarity is reflected in the decreasing total variation distance between the distributions of human and AI-generated text sequences. Adversaries, by seeking to mimic the human-generated text distribution using AI models, implicitly reduce the total variation distance between the two distributions to evade detection. Recent attack work shows that as the total variation between the two distributions decreases, the performance of even the best possible detector deteriorates~\cite{sadasivan2023can}. Paraphrasing as described above is only one mechanism for attackers to destroy watermarks, with emoji attacks~\cite{kirchenbauer2023watermark} (where the model is asked to rewrite the text with some transformation such as emoji-after-each-word applied) and related generative attacks posing new challenges to watermark detectors. Questions about the existence and feasibility of ``semantic'' watermarks, i.e., watermarks independent of the stylistic choices in the output, are still very much open to further research.

In conclusion, no provably robust watermark exists, and we need to resort to empirical evaluation which is prone to methodical errors. Existing watermarking algorithms only withstand attacks when the adversary has no access to the detection algorithm. There exist adaptive attacks to break known (human-made) watermarking algorithms. Moreover, it is an open question whether a deep neural network, if accessible by the adversary, can serve as a detection algorithm that is robust even in the presence of adversarial examples.

\chapter{Long-Term Goals}\label{sec:long}

Beyond the concrete issues raised in Section~\ref{sec:short} that we propose should be considered in the next couple of years, there are more fundamental challenges with the creation and use of GenAI systems. Addressing these will require combining deep, technical advances with novel social, political, cultural, and economic mechanisms.
\begin{itemize}
\item \textbf{Need for socio-technical solutions.}
GenAI is a groundbreaking technology with the potential to yield profound societal impact due to its advanced capabilities and widespread use. However, there is an inherent disconnect between the technical solution and the social requirements for its deployment in diverse contexts. Human activity is remarkably adaptable, nuanced, and context-driven, whereas computational algorithms often exhibit a ``rigid and brittle'' nature, perhaps inevitably due to their reliance on mechanisms such as formalization and abstraction~\cite{sociotechnical}. Going forward, it is imperative to comprehend the interplay between technology and society and make earnest efforts to bridge this socio-technical gap.

One critical aspect of this endeavor is the need for new model evaluation metrics. GenAI models operate in open-ended and complex output spaces, where determining what constitutes a ``good'' output can be multifaceted and context-dependent. Traditional model evaluation metrics, such as accuracy and performance, fall short of capturing the full scope of this complexity. Additionally, the diverse and uncertain capabilities of GenAI models, owing to their general-purpose nature, exacerbate the challenge. Consequently, there is a growing need for novel evaluation metrics that incorporate social awareness. This entails understanding the social requirements of downstream applications and designing customized metrics that explicitly articulate the drawbacks and trade-offs~\cite{liao2023rethinking}.

As we navigate through an ever-expanding online landscape, the ability to discern the trustworthiness of digital content becomes paramount. The question of central importance is:
\begin{quote}
Do I trust the person I am interacting with?
\end{quote}
To tackle this issue, an online reputation system can be developed. Within this system, users would be encouraged to establish and maintain a public digital identity with verifiable credentials. By linking this identity to various platforms and accounts across the web, a consistent online presence could be established. This practice is analgous to the traditional bylines of newspaper articles, where the name of the reporting journalist adds credibility to the content. Furthermore, the reputation system should be designed to track the chain of information dissemination. Such a system would empower users to trace the origin and subsequent sources of information, providing transparent insights into the content's journey from its original creator to its current state. We note that reputation-based mechanisms do not rule out the need for privacy in such online settings.

Accountability is another crucial aspect that needs to be modeled. Liability for harms that could have been averted through more careful development, testing, or standards could provide strong incentives for responsible practices~\cite{EU}. Individual users should be held responsible for deliberate misuse or negligence. Similarly, developers and providers of GenAI models should bear legal liability for the actions of the models. Further investigation is necessary to determine how accountability should be applied and liability assigned across users, model developers, and model providers~\cite{buiten_2023_ai-product-liability}.

Understanding the potential impacts on privacy in the era of GenAI is another critical yet complex task. Advanced GenAI models are trained on vast amounts of data scraped from the internet, including a wealth of personal information about individuals, ranging from personal preferences to potentially sensitive details. This raises severe privacy concerns, as training data can be extracted verbatim from the models in some cases~\cite{Carlini2020ExtractingTD}. While the data may be publicly available, that does not mean it was intended for utilization by third-party entities for commercial purposes.  This lack of explicit consent and the resulting unauthorized use of data introduces new dimensions of privacy concerns, as underscored by a recent lawsuit~\cite{lawsuit,copyrightlaw}. Formalizing the evolving notions of privacy that address the ethical implications of utilizing data in the public domain for model training poses a vital challenge.

An important point to note is that user attention is a limited resource in the digital ecosystem, and any solution should avoid overwhelming users. For instance, long and complex privacy policy agreements on websites and mobile applications are often not read by users. Simplicity and user-friendliness should be at the forefront of all solutions to ensure effective communication of important information.

\item \textbf{Multiple lines of defenses.} Given the complexity of tasks performed by state-of-the-art GenAI models, achieving perfect risk mitigation in safety solutions is an extremely challenging endeavor. Instead, we should adopt novel risk management strategies that allocate resources to areas of highest vulnerability and prioritize safety measures accordingly. To build a comprehensive GenAI safety strategy, multiple lines of defense are required.

The first line involves training-time interventions to align models with predefined values~\cite{ouyang2022training, bai2022constitutional}. A common approach is reinforcement learning from human feedback (RLHF)~\cite{stiennon2022learning}. LLMs  can be prompted to perform a range of NLP tasks. However, the language modeling objective used for training -- predicting the next token -- differs significantly from the objective of ``following the user's prompt helpfully and safely,'' which complicates the assessment of the quality of generated NLP text. Additionally, this evaluation is subjective and context-dependent, making it challenging to be captured via mathematical notions, such as a loss function. RLHF addresses this challenge by directly using human preferences as a reward signal for fine-tuning.

The next line of defense involves the post-hoc detection and filtering of inputs and outputs~\cite{gehman-etal-2020-realtoxicityprompts, solaiman2021process, welbl2021challenges, xu2021recipes} to catch inappropriate content that might slip through. Pre-training filtering may not capture all potential sources of bias or harmful content, especially as GenAI models encounter diverse and evolving data sources. Post-hoc detection helps identify any unforeseen biases or harmful patterns that may emerge during the model's usage. An additional advantage is customizability---post-hoc detection can be tailored to the specific needs of different applications, ensuring a more personalized and context-aware approach to content moderation. Furthermore, post-hoc detection aids in reducing false positives, ensuring legitimate content is not unnecessarily restricted. It is important to take into account the theoretical impossibility of detection and filtering and consider how ML techniques may be combined with security techniques~\cite{glukhov2023llm}.

Finally, the above efforts must be complemented by red teaming, which proactively identifies vulnerabilities, weaknesses, and potential blind spots~\cite{ganguli2022red,gpt4,Perez2022RedTL}.    Red teaming adopts an attacker's mindset and conducts rigorous stress testing. The goal is to simulate real-world attack scenarios and provide a practical assessment of the security measures in place by structurally probing the models. This step is especially critical for a consumer-facing technology like GenAI, which is accessible to a wide range of users and thus may potentially be targeted by a large pool of adversaries. Ensuring that the red team comprises a diverse group of experts is crucial to maximizing the efficacy of this approach.

\item \textbf{Pluralistic value alignment.} Value alignment of AI refers to the crucial process of ensuring that AI functionalities are in harmony with human values and objectives. However, a key question in this regard revolves around determining to what or whose values AI systems ought to align with~\cite{Gabriel_2020}. In particular,  how do we decide which principles or objectives to encode in AI, and who holds the right to make these decisions? The complexity of this issue is exacerbated by the fact that we inhabit a pluralistic world with diverse and competing conceptions of values rooted in cultural, political, and individual beliefs. Furthermore, navigating legal jurisdiction presents additional challenges, particularly when dealing with multinational organizations operating across different legal frameworks. With machines displaying human-like qualities, intriguing new questions emerge. For example, freedom of speech has long been recognized as a fundamental and indispensable democratic right for humans. However, should we expect machines to be granted the same rights, or should we make a distinguishing case?

As we grapple with these complexities, it becomes evident that taking a binary stance on value alignment is problematic. Such an approach oversimplifies the nuanced spectrum of what is considered good or bad, and fails to capture the multifaceted nature of value systems in our ever-changing society. To address these challenges, it is essential to engage in interdisciplinary discussions and collaborations. Ethicists, AI researchers, legal experts, policymakers, and the wider public should be involved in inclusive dialogues to navigate the intricacies of value alignment in AI. By acknowledging and respecting the pluralistic nature of our world, we can strive for the responsible development of artificial intelligence that upholds fundamental human values while adapting to the evolving relationship between humans and machines.

\item \textbf{Reduce barrier-to-entry for GenAI research.}
The development and training of large-scale foundation models demand substantial computational power, which can currently be afforded only by organizations with significant financial capital. Consequently, major advancements and breakthroughs in GenAI models have predominantly occurred within the commercial sector, leading to a concentration of power among a few tech giants. This raises serious concerns about potential centralization of influence, where a handful of companies wield unprecedented control over data, information, and decision-making processes. The consequences range from limiting healthy competition to stifling innovation and raise ethical concerns about responsible AI use. Moreover, the profit-driven nature of the commercial sector may prioritize economic interests over the scaling of AI safety research, creating a mismatch with societal well-being.

Hence, it becomes imperative to democratize GenAI research and reduce barriers to entry for academia and smaller startups. Fostering research on more efficient training processes for GenAI models, with a concerted effort from the government, may help address this issue. Additionally, promoting access to open-source solutions enhances transparency and reliability. However, a caveat in this regard is that open-sourcing also increases access for adversaries and thus the potential for misuse. For instance, recent work~\cite{zou2023universal} shows how to carry out automated safety attacks on open-source LLM chatbots that surprisingly transfer also to closed-source chatbots, such as ChatGPT, Bard, Claude. Similarly, follow-up work to detect such safety attacks was also demonstrated on open-source models~\cite{alon2023detecting}.

Moving forward, we need a more inclusive and collaborative AI ecosystem to harness the potential benefits of GenAI while mitigating risks and safeguarding against unethical use.

\item \textbf{New partnerships among stakeholders.} Currently, policy decisions are often made without sufficient consideration of the rapidly evolving technology landscape. However, as the use of GenAI proliferates, this approach is no longer sustainable. Such a disconnect between policy-making and technological advancements can have serious consequences, potentially hindering the efforts of those striving to safeguard AI research and innovation. A recent example of this is the contentious conflict between academics and the government over the EU's new legislation on content filtering~\cite{Euronews}.

To address these challenges and create a conducive environment for responsible GenAI development, we urgently need new partnerships that bridge the gap between the triumvirate of government, academia, and the commercial sector. These partnerships should be built upon principles of collaboration, open communication, and mutual understanding. Effective policy-making in the GenAI era requires the active involvement of experts from academia, who possess in-depth knowledge of the technology and its potential implications. It also demands the insights and perspectives of the commercial sector, which actively develops and implements GenAI solutions. Furthermore, government representatives need to be proactive stakeholders, taking an integral role in shaping policies that foster innovation, ensure ethical AI use, and safeguard societal interests. Moving forward, a shared vision and concerted effort can lead to well-informed and balanced policies that embrace the opportunities while addressing the challenges of GenAI.

\item \textbf{Grounding.} LLMs are increasingly being used in cybersecurity contexts, such as threat intelligence. For example, imagine a security analyst who gathers all the information (e.g., reports, images, speech recordings) related to a threat scenario, and then asks an LLM-powered application for a summary of the threat scenario. A summary should not contain hallucinations or ``made up facts,'' and most crucially, the summary of the threat scenario should only depend on the relevant data uploaded by the analyst. US Department of Defense  has identified this as a major issue (see the recent IARPA BENGAL program
\url{https://www.iarpa.gov/research-programs/bengal}). Hallucinations can have devastating effects also in other high-stakes settings (e.g., healthcare). This is exactly the problem addressed by grounding.

Formally, grounding requires that the text generated by an LLM is attributable to an authoritative knowledge source. Here, ``attribution'' means that a generic human would agree that the text follows from the authoritative source~\cite{rashkin2022measuring}. The specific authoritative source may vary across applications. For instance, a health question-answering system may consider only a selected set of healthcare journals as authoritative.

There are two broad categories of works in the area of LLM grounding -- (1) Detecting whether a given LLM response is grounded, and (2) Encouraging LLMs to generate grounded responses. A popular approach for (1) is to use a separate natural language inference (NLI) model to test whether the generated text is entailed by the knowledge text. Other approaches include comparing the generated and knowledge texts using BLUERT~\cite{rashkin2022measuring}, BERTScore~\cite{zhang2020bertscore}, and other text similarity metrics. A recent evaluation finds the NLI-based approach to achieve strong results compared to the alternatives~\cite{honovich2022true}. In cases where the knowledge source is a corpus of documents rather than a single piece of text, an additional step is to retrieve the relevant knowledge text from the corpus. This is typically done by mapping the generated text to a fact-checking query (e.g., the response ``Joe Biden is the president of the United States'' is mapped to ``Who is the president of the United States?''), and using an off-the-shelf retrieval system to obtain the relevant knowledge text.

There have been several works exploring how to make LLMs generate grounded responses in the first place. A simple and effective method is to augment the prompt with relevant knowledge snippets, as well as additional instructions asking the LLM to only use information available in the provided snippets; see~\cite{ram2023incontext} for extensions of this idea. This approach is attractive as it only requires API access to the LLM. Other approaches involve tuning the LLM to generate grounded responses with relevant citations. One approach is to tune the LLM's weights on a dataset of query and (grounded) response pairs. Another approach is to use reinforcement learning to tune the weights based on feedback on groundedness and plausibility of generated responses~\cite{menick2022teaching}. Such feedback may be obtained by training a reward model on human ratings, or using an NLI-based grounding detection model discussed above.

Finally, a third strand of work involves iteratively revising an LLM's response in cases where it is found to be ungrounded. A common approach here is to prompt the LLM back with feedback on how grounding fails, as shown in~\cite{gao2023rarr}:
\begin{quote}
\texttt{You said: \{text\}, I checked: \{query\}, I found this article: \{knowledge\}, This suggests $\ldots$}
\end{quote}

The three strands of work discussed above make some headway on solving the problem of grounding for LLMs. However, there are still plenty of open questions. On the detection side, there is a continuum between blatant hallucinations and fully grounded responses. For instance, a response on the side effects of a drug may be grounded but biased, in that it does not convey all opinions mentioned in the knowledge corpus. Natural language entailment-based checks cannot detect such bias. On prompting models to generate grounded responses, a number of prompting strategies~\cite{gao2023rarr,madaan2023selfrefine,yao2023react} have been proposed. However, designing these strategies is still an art, and there is no clear consensus on which strategy works best. When tuning models to generate grounded responses, a common pitfall is that the model loses its creativity, and resorts to quoting verbatim from the knowledge sources. Avoiding this requires carefully balancing various training objectives---fluency, grounding, plausibility, etc. Finally, more fundamental research is needed to understand why LLMs hallucinate (especially nonsensical text) and to design training strategies to mitigate such behavior.
\end{itemize}

\chapter{Conclusion}\label{sec:conclusion}

Every significant technological advancement, such as GenAI, surfaces the dual-use dilemma. Our work investigates the attack and defense capabilities of these powerful technologies. {\it This work is not meant to be the ``final word'' on this topic.} Our work has been shaped by the workshop held at Google, but that also means that the viewpoint is biased toward the speakers and attendees of the workshop. Short-term goals (Section~\ref{sec:short}) describe some problems that the community should start immediately investigating. Long-term goals (Section~\ref{sec:long})  describe problems that are challenging and thus will require significant research effort. This workshop is just a start, and we believe we need several such meetings to explore the complete landscape. However, we believe that an investigation into GenAI risks and their mitigation is crucial and timely, especially since attackers have already started using these technologies, and defenders must not be caught ``flat-footed.''

\begin{acknowledgements}

We thank the anonymous reviewers together with Jindong Wang (Microsoft), Anthony Gitter (University of Wisconsin, Madison), Gabriel Alon (University of Michigan), Ashwinee Panda (Princeton University), Alejo Jos\'e Sison Galsim (Universidad de Navarra), and Jonas A. Geiping (University of Maryland, College Park) for their insightful feedback.\vspace*{6pt}

\noindent {\bf Special note:} This document was cleared by DARPA on August 28, 2023. $^{**}$ All copies should carry Distribution Statement ``A'' (Approved for Public Release, Distribution Unlimited). If you have any questions, please contact the Public Release Center.

\end{acknowledgements}

\setcounter{biburlnumpenalty}{600}
\setcounter{biburlucpenalty}{600}
\setcounter{biburllcpenalty}{600}

\printbibliography

\end{document}